\documentclass{llncs}

\usepackage{makeidx}
\usepackage{graphicx}

\begin{document}



\mainmatter

\title{Object classification in images of Neoclassical furniture using Deep Learning}


\author{Bernhard Bermeitinger \and Andr\'e Freitas \and Simon Donig \and Siegfried Handschuh}
\authorrunning{Bernhard Bermeitinger}


\institute{
	Universit\"{a}t Passau, 94032 Passau, Germany\\
	\email{bernhard.bermeitinger@uni-passau.de}}

\maketitle


\section{Introduction}\label{sec:Introduction}
	This short paper outlines research results on object classification in images of Neoclassical furniture. The motivation was to provide an object recognition framework which is able to support the alignment of furniture images with a symbolic level model. A data-driven bottom-up research routine in the \emph{Neoclassica} research framework is the main use-case.
	This research framework is described more extensively by Donig et al.~\cite{Donig2016}. It strives to deliver tools for analyzing the spread of aesthetic forms which are considered as a cultural transfer process.

	To robustly identify artifacts that are shaped in accordance particular aesthetic programs (such as Classical styles) is an important step on a way to being able, to actually distinguish between artifacts bearing the traits of different styles. Conventionally, historians of art have relied chiefly on cataloging and canonization as tools for analyzing changing aesthetic movements. The \emph{Neoclassica} framework seeks to provide them with new digital tools that help to process a broader range of sources from more diverse provenance by aggregating data provided by GLAM-partners  (Galleries, Libraries, Archives, and Museums).

	By introducing Deep Neural Network Models from Machine Learning (ML) to this field, we hope that in particular semi-supervised learning methods will uncover clusters that were previously unknown. \emph{Classification} in ML enables automatic detection of patterns and has recently found interest for instance in visual culture research. Replication of cognitive tasks such as the analysis of visual art has long been a challenge for computers. By employing supervised and unsupervised learning methods, machines have recently been able to create classifications (e.g.~by Shamir et at.~\cite{Shamir2012a}) for schools and influences among painters that show a remarkable resemblance to those of human experts.

	A particular promising field for applying ML seems repetitive features like aesthetic forms. That makes it especially suited for analyzing artistic styles like in the case of Shamir et al.~\cite{Shamir2012} but also material artifacts like architecture or furnishings consisting of such features as pointed out by Prown et al.~\cite{Prown1980}. Naturally, physical artifacts of past centuries are becoming rarer as time goes on. During the era of Neoclassicism in the 18th and 19th century, photography was not yet invented and then later very complicated in its early stages. This results in a very small number of images of instances of different types of artifacts. Most deep learning approaches are trained on millions of images. At the time of the experiments, the \emph{Neoclassica} framework contained roughly 2\,000 images; three orders of magnitude less than other methods.
	By using Neural Networks for ML the task at hand is a \emph{Multi-Label Classification} (MLC) task. Applying \emph{pre-training} substantially reduces the amount of images required for higher accuracy.

\section{Experiment}\label{sec:Experiment}
   
    \subsubsection{The \emph{Neoclassica} data set:}
	    To train the Neural Networks for the tasks at hand we compiled a data set of the most common furniture types of the Neoclassic era from 1770 to 1840. It consists of 2\,167 RGB-encoded images in 300 categories. Each category represents one \emph{artifact}. An artifact is a combination of different labels that occur multiple times throughout the data set. The most artifacts are represented by 4--9 images with 6 being the median. First experiments use a \emph{Multi-Class Classification} task, so the 300 artifacts are reduced to 42 classes with 52 images per class on average.

	    The proposed approach makes use of the current superior accuracy (as in the \emph{ImageNet Large Scale Visual Recognition Challenge} by Russakovsky et al.~\cite{Russakovsky2015}) of \emph{Convolutional Neural Networks} (CNN) in image classification tasks to recognize artifacts from the era of \emph{Neoclassicism}.
	    
	    The layout of the CNN is a custom implementation of \emph{VGG-16} by Simonyan et al.~\cite{Simonyan2014}. It consists of five groups of convolutional layers followed by a max-pooling layer. Two fully connected layer with each 4096 nodes are the last layers before the output units. They are connected with a dropout layer. Each convolutional layer and the two fully connected layers have \emph{ReLU} (recified linear unit) as activation function. The output layer has the required number of units for the current task and the \emph{softmax} function as activation function.
    
    \subsubsection{Notes on pre-training:}
    	Pre-training is a commonly used method for improving the overall performance of a Neural Network. The assumption is that by training the classifier on many images that are structured with annotations the classifier trains basic features like edges and color differences and stylistic features like round shapes and specific edges. These extracted features are exploited in the unknown data set and immediately result in better performance from the beginning.

    \subsubsection{Notes on image augmentation:}
    	Image augmentation is a common procedure during training to virtually increase the number of available images. There are many different possibilities to change the appearance of images: rotation, translation, flipping, and cropping are four examples.
    	Randomly cropping the image to smaller patches and randomly flipping the image on the vertical axis are two methods that are applied in this experiment.

    \subsubsection{Results:}
        The experiments are run sequentially on a dedicated server with two Intel Xeon E5-2637 processors running at 3.6 GHz, 64 GB of RAM and one NVIDIA Tesla K40c graphics card with 12 GB of memory. The experiments are implemented in Python 3.4 with \emph{Lasagne} \cite{lasagne} and \emph{Theano} \cite{theano} as main neural network and computational frameworks.
        
    	The first experiment is done without pre-training, hence using only the \emph{Neoclassica} data set on a newly initialized CNN. Throughout the experiments the following parameters stay the same: The image size is always $ 120 \times 120 $, the batch size is set to 256, the learning rate is set to 0.03 and the momentum is set to 0.9. The loss is computed with the \emph{categorical crossentropy} loss function. The main experiments are each separated into four different configuration setups: using colored images (\texttt{3\_}) or grayscale (\texttt{1\_}) and using augmentation during training (\texttt{\_Y}) or not (\texttt{\_N}). A random train/test split of 80/20 is done prior to running the experiment. Each configuration is trained and validated on the same image set. After pre-training, the weights for the layers of the pre-trained network are exported and imported into a newly instantiated network with the same layout. Only the number of output units is adjusted accordingly to match the 42 classes of the experimental \emph{Neoclassica} data set.

	    \begin{table}
	    \centering
	    \caption{Comparison of F1-measures between different configurations with and without pre-training}
	    \label{tbl:F1results}
	    \begin{tabular}{c|c|rrrr}
	    	         &   ~config  & ~on pre-training & ~no pre-training & ~with pre-training & ~improvement \\ \hline
	    	   F1    & \texttt{1N} &             0.478 &             0.330 &               0.347 &            5\% \\
	       	         & \texttt{1Y} &             0.533 &             0.320 &               0.400 &           25\% \\
	    	         & \texttt{3N} &             0.453 &             0.206 &               0.369 &           79\% \\
	    	         & \texttt{3Y} &             0.543 &             0.333 &      \textbf{0.442} &           32\% \\
	    	         \hline
        	Accuracy & \texttt{1N} &             0.457 &             0.323 &               0.407 &           26\% \\
        	         & \texttt{1Y} &             0.528 &             0.322 &               0.409 &           27\% \\
        	         & \texttt{3N} &             0.450 &             0.218 &               0.416 &           91\% \\
        	         & \texttt{3Y} &             0.539 &             0.368 &      \textbf{0.438} &           19\% \\
        	         \hline
	    \end{tabular}
	    \end{table}
	    
	    As Table~\ref{tbl:F1results} shows, a pre-training step leads to a consistent improvement in F1-measure and accuracy. On average, the F1-measure is improved by 35.25\%, the accuracy by 41.75\%. These high average values are tainted by one configuration: \texttt{3N} (using colored images without augmentation), meaning that pre-training provides a high F1-measure and accuracy improvement on colored images without augmentation. This leads to the observation that the application of pre-training always gives better performance, especially without the application of augmentation. Applying augmentation to the training images also improves the F1-measure but it is already higher, so the improvement rate is lower but the overall result is still superior.

	    Additionally, some classes in the data set are not mutually exclusive. For example, there is a constructional difference between ``armoires'' and ``secretaries''. But their similarity score is very high, as opposed to for example their individual similarity scores to ``beds''. In the range of different types of furniture ``armoires'' and ``secretaries'' are too close to each other such that their numerical similarity is too small to make a noticeable difference.

\section{Conclusion}\label{sec:Conclusion}
	The approach in this work evaluated a \emph{Convolutional Neural Network}, namely \emph{VGG-16} by Simonyan et al.~\cite{Simonyan2014} for a Multi-Class Classification task within the domain of furniture recognition. The implementation was applied to the custom data set \emph{Neoclassica} specifically adapted for this work. There were four different network configurations examined: using grayscale or RGB images and apply augmentation during training or not. The experiments show that using augmentation techniques always lead to a higher F1-measure (26\% on average).
	Moreover, another experiment examined the application of pre-training with manually selected images from ImageNet that roughly match the classes from the \emph{Neoclassica} data set. Pre-training improved the F1-measure for all four configurations by 31\% from 0.297 to 0.390 on average. The highest averaged F1-measure overall \emph{Neoclassica} classes of 0.442 is achieved by using RGB images and augmentation during training. The highest average accuracy is also achieved by this configuration and yields a success rate of 43.8\%.

\bibliographystyle{splncs03}
\bibliography{references}


\end{document}